\documentclass[10pt,twocolumn,letterpaper]{article}

\usepackage[accsupp]{axessibility}  
\usepackage{iccv}              
\usepackage{multirow}
\usepackage[hyphens]{url}
\usepackage{bm}
\usepackage{amsmath}
\usepackage{graphicx}
\usepackage{pifont}
\usepackage{booktabs}
\usepackage[normalem]{ulem}
\definecolor{iccvblue}{rgb}{0.21,0.49,0.74}
\usepackage[pagebackref,breaklinks,colorlinks,allcolors=iccvblue]{hyperref}

\def\paperID{XXXX} 
\def\confName{ICCV}
\def\confYear{2025}

\title{Lattice-allocated Real-time Line Segment Feature Detection \\and Tracking Using Only an Event-based Camera}

\author{Mikihiro Ikura\\
Istituto Italiano di Tecnologia\\
{\tt\small mikihiro.ikura@iit.it}
\and
Arren Glover\\
Istituto Italiano di Tecnologia\\
{\tt\small arren.glover@iit.it}
\and
Masayoshi Mizuno\\
Sony Interactive Entertainment Inc.\\
{\tt\small masayoshi.mizuno@sony.com}
\and
Chiara Bartolozzi\\
Istituto Italiano di Tecnologia\\
{\tt\small chiara.bartolozzi@iit.it}
}

\begin{document}
\maketitle
\begin{abstract}
Line segment extraction is effective for capturing geometric features of human-made environments. Event-based cameras, which asynchronously respond to contrast changes along edges, enable efficient extraction by reducing redundant data. However, recent methods often rely on additional frame cameras or struggle with high event rates.
This research addresses real-time line segment detection and tracking using only a modern, high-resolution (i.e., high event rate) event-based camera. Our lattice-allocated pipeline consists of (i) velocity-invariant event representation, (ii) line segment detection based on a fitting score, (iii) and line segment tracking by perturbating endpoints. Evaluation using ad-hoc recorded dataset and public datasets demonstrates real-time performance and higher accuracy compared to state-of-the-art event-only and event-frame hybrid baselines, enabling fully stand-alone event camera operation in real-world settings.
\end{abstract}    
\section{Introduction}
\label{sec:line-track:intro}
Line segment on images can efficiently encode geometric features of human-made structures, aiding to reduce computational cost for crucial vision tasks such as 3D reconstructions~\cite{Denis2008}, Structure from Motion (SfM)~\cite{Micusik2014}, Simultaneous Localization and Mapping (SLAM)~\cite{Qiao2021}, and pose estimation~\cite{Xu2017}. Recently, compared to classic computer vision approaches, such as the Line Segment Detector (LSD)~\cite{Grompone2012}, line segment extraction methods with deep learning approaches~\cite{Pautrat2023, Gu2022} have become more widespread. These deep networks have been trained with particularly challenging images to ensure highly accurate and robust performance. However, the frame input to the network still suffers from quality degradation due to motion blur for fast moving cameras or targets, and pixel intensity saturation in high dynamic range (HDR) conditions.
Moreover, their high computational demands that usually require the use of GPUs limit use on resource-constrained platforms such as embedded systems and mobile robots.

\begin{figure}[t]
    \centering
    \includegraphics[width=1.0\linewidth]{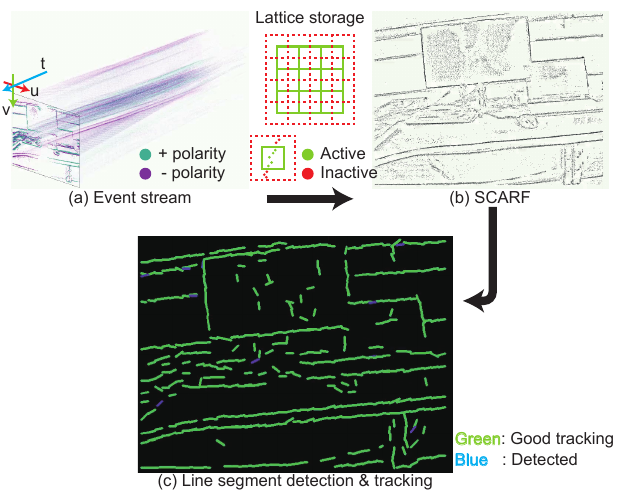}
    \caption{The proposed lattice-allocated real-time line segment detection and tracking. The (a) events are stored into lattice storage by (b) a velocity-invariant method SCARF. Line segments are (c) initialized within a block using only events from the active region of the lattice storage (Detection). Subsequently, their positions are updated both inside and outside the block using events from the active and inactive regions (Tracking).}
    \label{fig:line-track:introfigure}
\end{figure}

Event-based cameras, inspired by biological vision, are gaining popularity in robotics and computer vision for low-latency, high dynamic range, and robustness to motion blur~\cite{Gallego2022, Bharatesh2024}. They asynchronously capture pixel-wise brightness changes as events $e = \{t, u, v, p\}$, where $t$ is the timestamp with microsecond resolution, $(u,v)$ the pixel location, and $p \in \{1, -1\}$ the polarity indicating the direction of the brightness change. This allows edge extraction as event clusters (Fig.~\ref{fig:line-track:introfigure}~(a)) and has motivated line segment detection and tracking methods~\cite{Grompone2012, Everding2018, Dietsche2021}. Recent event cameras achieve high resolutions (e.g., 640×480, 1280×720), less noise, and event rates exceeding 1 Giga events/s, surpassing older models like DAVIS346 (346×260, 12 Mev/s). However, this brings challenges for real-time, event-by-event processing. In addition, while event-frame hybrid methods~\cite{Huai2023, Wang2024, Choi2025} have improved performance, their reliance on frame cameras limits the potential of fully event-based processing despite recent hardware advancements of event cameras.

\renewcommand{\thefootnote}{\fnsymbol{footnote}}\setcounter{footnote}{1}
To address the challenges of real-time and low computational cost line segment detection under high event rates, we propose a novel model-based pipeline (Fig.~\ref{fig:line-track:introfigure}) that (i) stores events using a velocity-invariant method, ``SCARF'' (A \underline{\textbf{S}}et of \underline{\textbf{C}}entre \underline{\textbf{A}}ctive \underline{\textbf{R}}eceptive \underline{\textbf{F}}ields), (ii) detects line segments based on a fitting score from a lattice, and (iii) tracks them by perturbating their endpoints inside and outside the lattice. SCARF efficiently handles large event volumes compared to prior methods~\cite{manderscheid2019speed, glover2021luvharris, glover2024edopt}, enabling real-time operation on high-resolution cameras (e.g., 640×480). By leveraging events stored in the preallocated lattice storage as a batch, our method achieves over 200 Hz for detection and 400 Hz for tracking in a process-driven manner. Once detected, line segments are passed to the tracker, reducing computational cost and increasing efficiency through cooperative processing. Overall, our method enables real-time performance at high event rates and dense short-segment extraction for complex shapes with high accuracy.

In summary, the contributions of this paper are:
\begin{itemize}
    \item Introduction of lattice-allocated real-time pipeline that, for a recorded dataset (640x480), achieves event-driven storage at over 20 Mev/s, process-driven line segment detection at over 200 Hz, and tracking at over 400 Hz.
    \item Qualitative and quantitative evaluations with locally recorded dataset (640x480) and a publicly available datasets (240x180, 346x260) for line segment accuracy, real-time performance, and parameter sensitivity
    \item C++ implementation of the proposed pipeline and recorded dataset are released as open-source: \url{https://github.com/event-driven-robotics/RT-EvLDT}.
\end{itemize}

\section{Event-driven Line Segments Detection and Tracking Methods}
\label{sec:line-track:related_work}
Event cameras naturally emphasize edges, therefore, event-based line feature detection and tracking was utilized for many computer vision and robotics tasks, such as 6DoF pose tracking~\cite{Zibin2024, Zibin2025}, robot control~\cite{Dimitrova2020, Gómez2020}, visual odometry~\cite{Gentil2020}, and SLAM~\cite{Chamorro2022}. However, the focus of these applications does not consider computational cost scaling with camera speed and resolution, leading to non-real-time process.~\cite{Zibin2024, Zibin2025} present iterative optimizations by minimizing error to improve accuracy. In addition, robot control methods~\cite{Dimitrova2020, Gómez2020} suffer from heavy computational cost by event-by-event processing. Furthermore,~\cite{Chamorro2022} contributes to line SLAM with light-weight line segment extraction based on hough-transform to avoid time-consuming line clustering such as IDOL~\cite{Gentil2020}.

According to~\cite{Dietsche2021}, event-based line detection and tracking methods were typically categorized into: (i) Hough-transform based, (ii) Non-parametric, and (iii) Spatio-temporal approaches.
Hough-transform based methods~\cite{Dimitrova2020, Gómez2020} find lines that contain the largest number of events in a line-based parameter space, but are computationally intensive and unsuitable for high event rates.
Non-parametric methods~\cite{Brändli2016} cluster pixels based on spatial derivatives from a time surface~\cite{Benosman2012, Mueggler17BMVC}. While line segments can be extracted even for complex shapes, detected segments are often short and sensitive to pixel-level noise.
Spatio-temporal methods assume that events aligned with a line within a sufficiently small time window lie on a plane in ${u,v,t}$ space, enabling line extraction via plane fitting~\cite{Everding2018, Dimitrova2020}. This yields longer, more stable segments, though performance on detailed or curvy shapes is limited.
Beyond these, deep learning approaches combine events and frames~\cite{Wang2024, Huai2023}, and contrast maximization~\cite{gallego2018unifying} enables event-image sharpening via motion compensation, leading to accurate line segment extraction~\cite{Choi2025}.
Our method takes a pseudo-spatio-temporal approach: events are buffered in a lattice structure following the FIFO principle without precise timestamps (pseudo-temporal); line fitting is performed independently within each block in ${u,v}$ space (pseudo-spatial).

\section{Lattice-allocated Real-time Line Segment Detection and Tracking}
\label{sec:proposed_method}
\begin{figure}[t]
    \centering
    \includegraphics[width=1.0\linewidth]{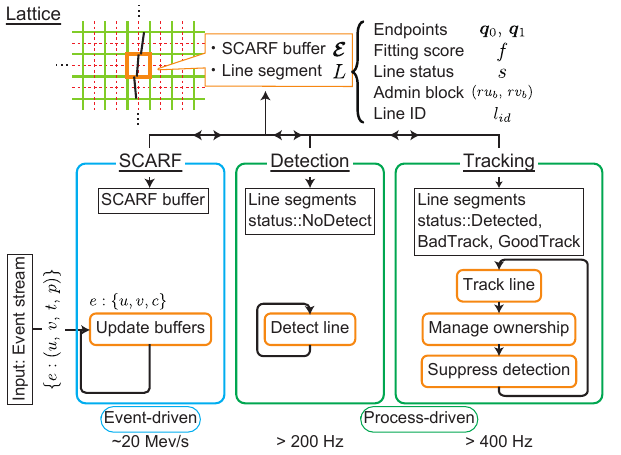}
    \caption{Parallelized pipeline overview of lattice-allocated line segment detection and tracking}
    \label{fig:line-track:3_overview}
\end{figure}
Figure~\ref{fig:line-track:3_overview} illustrates the parallelized pipeline proposed for lattice-allocated real-time line segment detection and tracking with event-based cameras. It mainly consists of three threads: SCARF, Detection, and Tracking. Each predefined grid block from a lattice consists of a line segment $L$ and a SCARF ring buffer $\bm{\mathcal{E}}$. The line segment $L$ is defined by: (i) two endpoints $\bm{q}_0 = (u_{q_0}, v_{q_0})$, $\bm{q}_1 = (u_{q_1}, v_{q_1})$; (ii) a fitting score $f$, computed during detection and tracking; (iii) a Line status $s \in \{$NoDetect, Detected, ProhibitDetection, BadTrack, GoodTrack$\}$; (iv) the block coordinate that owns the line segment, referred to as the Admin block $(ru_b, rv_b)$; and (v) a line segment ID $l_{id}$.

Firstly, in the SCARF thread, input from the event-based camera is processed event-by-event and is stored the SCARF buffers with respect to the event coordinate. By utilizing the stored events, the detection and tracking threads are not required to operate in an event-driven manner, allowing for the construction of a process-driven pipeline. The separation of event- and process-driven threads mitigates the difficulty of real-time processing that arise with increasing input event rates. 

In the detection thread, if no line segment currently exists in a block (i.e. $\text{LineStatus} == \text{NoDetect}$), a new line segment is extracted from the block using events stored in the SCARF buffer. 

In the tracking thread, the state of a line segment is continuously updated by slightly perturbating its two endpoints $\bm{q}_0, \bm{q}_1$ from the previous detection or tracking loop (i.e. $\text{LineStatus} \in \{ \text{Detected}, \text{GoodTrack} \}$). Additionally, this thread manages the suppression of detections in neighboring blocks to avoid conflicts between detection and tracking (i.e. $\text{LineStatus} == \text{ProhibitDetection}$) while ensuring that the owner of each line segment is consistently updated. In addition, if the fitting score $f$ falls below a predefined threshold $f_{th}$ or the line segment moves out of camera's field of view, the line status is shifted to BadTrack. The next iteration of the detection thread can then attempt re-detection of a new line segment. 

Compared to the line detection process, the tracking process with perturbation is computationally more efficient. Therefore, the cooperation between detection and tracking reduces computational costs and enhances the process frequency in process-driven threads. The following subsections explains each thread in more detail.

\subsection{SCARF}
\begin{figure}[t]
    \centering
    \includegraphics[width=1.0\linewidth]{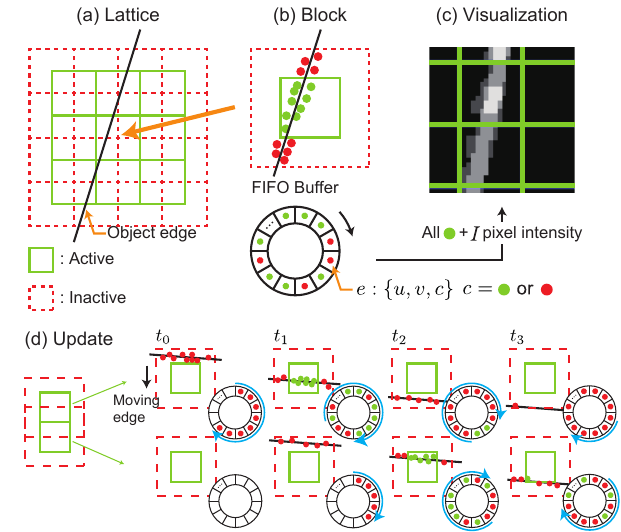}
    \caption{Overview of SCARF to store and visualize events with a lattice structure. (a) The lattice structure is defined with active and inactive regions. (b) Each block maintains a preallocated FIFO buffer for storing events. (c) For visualization, an image-like representation can be generated by assigning a pixel intensity $I$ to all ``active'' events within the buffer. (d) SCARF buffer update process. At $t_0$, the upper buffer is filled with inactive events. At $t_1$, it is ``excited'' by active events, while the bottom buffer is ``inhibited''. When the edge leaves the upper active region ($t_2$), the upper buffer becomes inhibited and the bottom buffer is excited. At $t_3$, the upper buffer is fully inactive, and inhibition begins in the bottom buffer.}
    \label{fig:line-track:3_SCARF}
\end{figure}
SCARF generates a consistent frame-like representation for scenes containing multiple objects moving at different speed.
SCARF adopts an approach where, instead of managing events with a buffer across the entire field of view (FOV), each predefined small block maintains its own buffer to store events. Such an approach assumes the motion within a single block is consistent over the block (e.g. not two objects moving with different speeds within a block) which becomes more valid as the block size decreases. Multiple objects, and camera motion, with different speeds are instead invariantly represented across different blocks.

Figure~\ref{fig:line-track:3_SCARF} shows that each SCARF block consists of an active region that covers the entire block and an inactive region, which is half the size of the block on each side and overlaps with adjacent blocks. Each event $e = \{t, u, v, p\}$ is therefore stored as an active event in the single buffer, and as an inactive event in up to three neighboring buffers. Each buffer has a finite size $N = \alpha \times b^{2}$, where $\alpha$ is a tunable parameter and $b$ is the block size [px], and is managed by FIFO (First-In, First-Out) principle, whereby the newest event replaces the oldest event in the buffer. Events are removed after N subsequent (active/inactive) events have been added to the buffer. Block overlap and inactive events are very important and are used to ``clear'' blocks as edges move out of the block region. They do so by overwriting active events, and are not considered themselves by any downstream visualization or processing, as shown in Fig.~\ref{fig:line-track:3_SCARF}. In this study, the raw events stored in the buffer are used for line segment detection and tracking. In addition, all active events stored in all blocks can be visualized on the image plane by accumulating a pixel intensity $I$ for each event at its pixel location $\{u,v\}$.

\subsection{Detection}
\begin{figure}[t]
    \centering
    \includegraphics[width=1.0\linewidth]{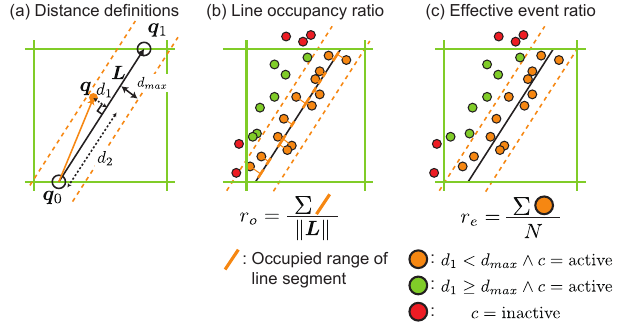}
    \caption{Definition of fitting score}
    \label{fig:line-track:3_fitting_score}
\end{figure}
This study unifies SCARF active blocks and the blocks for line segment management. The main objective of detection is to initialize line segments consisting of endpoints $\bm{q}_0, \bm{q}_1$ and fitting score $f$ inside blocks. An optimal line is fitted to the 2D ``active'' event point cloud extracted from the buffer in the block, with its intersections with the block boundaries defining the segment’s endpoints. This fitting process, which minimizes the total distance between the line and events, is computationally expensive, highlighting the importance of the lower-cost tracking process.

The fitting score $f$ is initialized with two ratios, (i) line occupancy ratio $r_o$ and (ii) effective event ratio $r_e$ shown in Fig.~\ref{fig:line-track:3_fitting_score} by following this equation:
\begin{equation}
    f = r_o \times r_e.
    \label{eq:line-trck:fittingscore}
\end{equation}
To compute these two ratios, the distances $d_1$ and $d_2$ shown in Fig.~\ref{fig:line-track:3_fitting_score} are defined as follows:
\begin{equation}    
    d_1 = \frac{\| (\bm{q} - \bm{q}_0) \times \bm{L} \|}{\| \bm{L} \|}, d_2 = \frac{(\bm{q} - \bm{q}_0) \cdot \bm{L}}{\| \bm{L} \|}
    \label{eq:line-track:d1d2}
\end{equation}
where $\bm{q}$ is the 2D position of an event $(u, v)$ from a buffer $\bm{\mathcal{E}}$, $\bm{q}_0$ is one of the endpoints of a line segment, and $\bm{L}$ is a vector on the line segment $\bm{q}_0 - \bm{q}_1$. $d_1$ represents the perpendicular distance from the point to the line, while $d_2$ denotes the distance of the projected point along the line segment. The line occupancy ratio quantifies how evenly events are distributed along the entire line segment. To calculate this, the occupancy token $T(i)$ is determined with:
\begin{equation}
T(i) =
\begin{cases}
1.0, & \text{if } i = \lfloor d_2 \rfloor \,\,\text{and}\,\, d_1 < d_{max}  \, \text{ for } \bm{q} \in \bm{\mathcal{E}}
 \\
0, & \text{else} \quad (0 \leq i \leq \lfloor \sqrt{2} b \rfloor ),
\end{cases}
\label{eq:line-track:occpancy_token}
\end{equation}
where $\lfloor \cdot \rfloor$ is a floor function, $d_{max}$ is a predetermined threshold parameter, and $b$ is block size. Finally, the line occupancy ratio $r_o$ is defined as:
\begin{equation}
    r_o = \frac{1}{\| \bm{L} \|} \sum_{i=0}^{\lfloor \sqrt{2}b\rfloor} T(i), \,(0 \leq r_0 \leq 1 ).
\end{equation}
\newcommand{\1}{\mbox{1}\hspace{-0.25em}\mbox{l}}
Effective event ratio $r_e$ can be calculated by counting ``active'' events closer to the line segment than $d_{max}$ from:
\begin{equation}
   r_e = \frac{1}{N}\sum_{p \in \mathcal{E}} \1(d_1 < d_{max} \land c = \text{active}), \, (0 \leq r_e \leq 1 )
   \label{eq:line-track:3_effectiveratio}
\end{equation}
where $c$ is a SCARF indicator, $N$ is the buffer size, and $\1(\cdot)$ is an indicator function.
Finally, the fitting score of the line segment $f$ is calculated with Eq.~(\ref{eq:line-trck:fittingscore}). The detection succeeds if this score is above the predetermined threshold $f_{th}$ and the detected line segment is switched to tracking process for continuous status updates.

\subsection{Tracking}
\begin{figure}[t]
    \centering
    \includegraphics[width=1.0\linewidth]{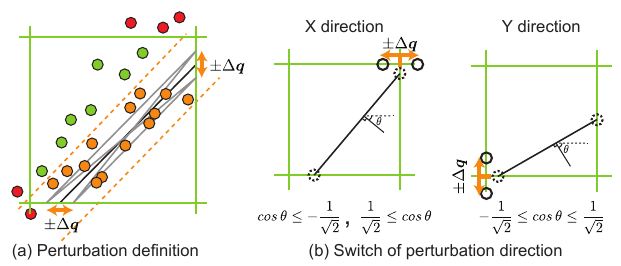}
    \caption{Perturbation of endpoints for tracking}
    \label{fig:line-track:3_tracking}
\end{figure}
Considering the assumption in SCARF that each line segment is approximately the size of the predefined block, endpoints of line segments can be fixed on lattice. This constraint for perturbation simplifies the tracking algorithm. The tracking algorithm in this research is inspired by~\cite{alzugaray2020, Ikura2024}, whereby hypothetical states $\mathcal{H}(\mathbf{x})$ are generated by perturbating tracking state $\mathbf{x}^{(k)}$ and the hypothetical state that produces the highest score is selected as the next tracking state $\mathbf{x}^{(k+1)}$. In this research, the tracking state is defined with the endpoints of the line segment as $\mathbf{x}^{(k)} = \{ \bm{q}_0,\, \bm{q}_1 \}$, and the tracking state is updated according to Eq.~(\ref{eq:line-track:3_trackupdate}), which consists of the fitting score $f$ of the line segment $L$ and SCARF buffer(s) $\mathcal{E}$:
\begin{equation}
\scalebox{0.85}{$
    \mathbf{x}^{(k+1)} =
    \begin{cases} 
    \underset{\bm{h} \in \mathcal{H}\left(\mathbf{x}^{(k)}\right)}{\arg \max }\,\, f\left(\displaystyle \bigcup_{(i,j) \in \mathcal{I}(L)} \mathcal{E}_a^{(k+1)}, \bm{h}\right), & \text{if } |\mathcal{I}(L)| > 1 \\[15pt]
    \underset{\bm{h} \in \mathcal{H}\left(\mathbf{x}^{(k)}\right)}{\arg \max }\,\, f\left( \mathcal{E}_{a \cup i}^{(k+1)}, \bm{h}\right), & \text{else} 
    \end{cases}
    $}
    \label{eq:line-track:3_trackupdate}
\end{equation}
where
\begin{equation}
    \mathcal{H}(\mathbf{x})=\{(\mathbf{x}=\{\bm{q}_0,\, \bm{q}_1\}) \cup\{\bm{q}_0 \pm \Delta \bm{q},\, \bm{q}_1 \pm \Delta \bm{q}\}\},
    \label{eq:line-track:3_hypotheticalstate}
\end{equation}
and $\mathcal{I}(L)$ represents coordinate of all blocks through which the line segment $L$ passes. Eq.~(\ref{eq:line-track:3_trackupdate}) considers two key factors: (i) A line segment spanning multiple blocks uses all active events $\bigcup_{(i,j) \in \mathcal{I}(L)} \mathcal{E}_a$ from their buffers, avoiding any redundancy due to overlap in the inactive region. (ii) A line segment within a single block considers both active and inactive events $\mathcal{E}_{a \cup i}$ for perturbations both within and beyond the block. The norm of the perturbation vector $\Delta \bm{q}$ is a tunable parameter, but the direction of $\Delta \bm{q}$ is determined by which direction (X/Y) each endpoint currently lies on in the lattice as shown in Fig.~\ref{fig:line-track:3_tracking}~(a). Therefore, the total number of fitting scores $f\left(\mathcal{E}, \bm{h}\right)$ to be evaluated is 5 based on the hypothetical state Eq.~(\ref{eq:line-track:3_hypotheticalstate}). When an endpoint gets sufficiently close to a block corner, the perturbation direction is adjusted to align more closely with the normal vector of the current line segment, as illustrated in Fig.~\ref{fig:line-track:3_tracking}~(b). This process prevents the line segment from extending infinitely while allowing perturbation both within and beyond the block, resulting in longer-lived line segments. After updating the tracking status, the best fitting score $f$ is compared with the same threshold $f_{th}$ used in detection to categorize the line segment as either GoodTrack or BadTrack.

\begin{figure}[t]
    \centering
    \includegraphics[width=1.0\linewidth]{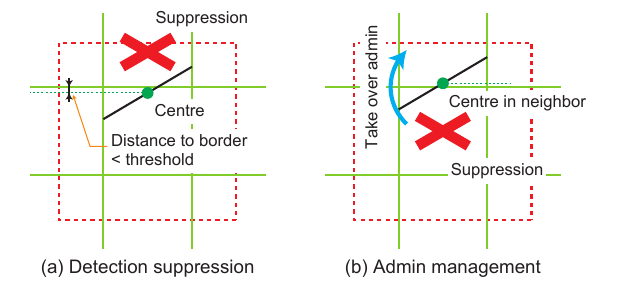}
    \caption{Detection suppression and admin management to avoid conflict between detection and tracking}
    \label{fig:line-track:3_prohibition}
\end{figure}
When a tracked line segment gets close to the border of the block, as shown in Fig.~\ref{fig:line-track:3_prohibition}, the detection and tracking are in conflict between 2 adjacent blocks. To avoid this conflict, the neighbouring block is suppressed to detect a new line segment when a centre of line segment is close enough to the block edge (Fig.~\ref{fig:line-track:3_prohibition}~(a)). Additionally, as the perturbation progresses and the centre of the line segment moves into a neighbouring block, the ownership (admin) of the line segment is transferred to that block and then the previous block is suppressed for detection (Fig.~\ref{fig:line-track:3_prohibition}~(b)). In the following processing step, that block will be responsible for tracking.
\section{Evaluation}

\subsection{Experimental Configuration}
\renewcommand{\thefootnote}{\arabic{footnote}}\setcounter{footnote}{1}
We compared the proposed line segment detection and tracking pipeline with four baselines: ELiSeD~\cite{Brändli2016}, Powerline~\cite{Dietsche2021}, C2F-EFIO~\cite{Choi2025}, and FE-LSD~\cite{Huai2023} as shown in Table~\ref{tab:line-track:4_requirements_baselines}. We modified and reimplemented the open source code$\footnote[1]{https://github.com/SensorsINI/jaer}$$^{,}$$\footnote[2]{https://github.com/uzh-rpg/line\_tracking\_with\_event\_cameras}$$^{,}$$\footnote[3]{https://github.com/choibottle/C2F-EFIO}$$^{,}$$\footnote[4]{https://github.com/lh9171338/FE-LSD}$ in C++ or Python environment for fair comparison. All methods were executed on a Intel(R) Core(TM) i7-9750H CPU @ 2.60 GHz x 12 Cores, 16 GB RAM at 2667 MT/s, and a GPU NVIDIA GeForce GTX 1650 for only deep-learning method FE-LSD. Further details about hyper-parameter settings in each baseline for evaluations are in the supplementary.
\begin{table}[t]
    \centering
    \scalebox{0.8}{
    \begin{tabular}{@{}ccccccc@{}}   
        \multirow{2}{*}{Baselines} & \multicolumn{3}{c}{Input} & \multirow{2}{*}{GPU} & \multicolumn{2}{c}{Line segments} \\
         & Event & Frame & IMU & & Detection & Tracking \\ \hline
        ELiSeD~\cite{Brändli2016} & \ding{51} & & & & \ding{51} & \ding{51} \\
        Powerline~\cite{Dietsche2021} & \ding{51} & & & & \ding{51} & \ding{51} \\
        C2F-EFIO~\cite{Choi2025} & \ding{51} & \ding{51} & \ding{51} & & \ding{51} & \ding{51} \\
        FE-LSD~\cite{Huai2023} & \ding{51} & \ding{51} & & \ding{51} & \ding{51} & \\
        Ours & \ding{51} & & & & \ding{51} & \ding{51} \\ \hline
    \end{tabular}
    }
    \caption{Comparison of input requirements, GPU dependency, and line segment outputs for ours and baseline methods}
    \label{tab:line-track:4_requirements_baselines}
\end{table}

The five pipelines were benchmarked using public datasets called Event Camera Dataset~\cite{Mueggler2017} (240x180) and MVSEC dataset~\cite{MVSEC2018} (346x260) which also include frames and arbitrary 6DoF motion from IMU. Additionally, to test the limits of real-time processing for higher event rates, we evaluated them on a novel dataset recorded from Prophesee's EVK3 (640x480). The quantitative evaluation on line segment accuracy requires the ground truth line segments. For recorded data with low complexity, ground truth line segments were created through hand annotation. For public datasets where pixel-to-pixel matching between RGB images and event data is ensured, LSD~\cite{Grompone2012} was applied to the RGB images to imitate the ground truth. Table~\ref{tab:line-track:4_datasets} describes the details of dataset for line segment evaluations.
\begin{table*}[t]
    \centering
    \scalebox{0.8}{
    \begin{tabular}{@{}cccccccc@{}}   
         & Dataset & Resolution & Complexity & Event & Frame & IMU & Line segment GT \\ \hline
        \multirow{3}{*}{\rotatebox{90}{\scriptsize Recorded}} & three\_vertical\_lines\_fast & 640x480 & Low & \ding{51} &  &  & Annotation \\
         & circle\_board & 640x480 & Low & \ding{51} & &  & Annotation \\
         & monitor\_6dof & 640x480 & High & \ding{51} & & & None \\ \hline
         \multirow{2}{*}{\rotatebox{90}{\scriptsize Public}} & ECD~\cite{Mueggler2017}: dynamic\_6dof & 240x180 & High & \ding{51} & \ding{51} & \ding{51} & LSD \\
         & MVSEC~\cite{MVSEC2018}: indoor\_flying1 & 346x260 & High & \ding{51} & \ding{51} & \ding{51} & LSD \\ \hline
    \end{tabular}
    }
    \caption{Datasets used for the evaluation of line segment detection and tracking. Each dataset has a duration of 10 seconds. For the MVSEC dataset in particular, the evaluation is performed over a time range from 15 to 25 seconds after the drone begins flying.}
    \label{tab:line-track:4_datasets}
\end{table*}

\subsection{Qualitative Evaluation}
\label{sec:line-track:qualitative_eval}
\begin{figure*}[t]
    \centering
    \includegraphics[width=1.0\linewidth]{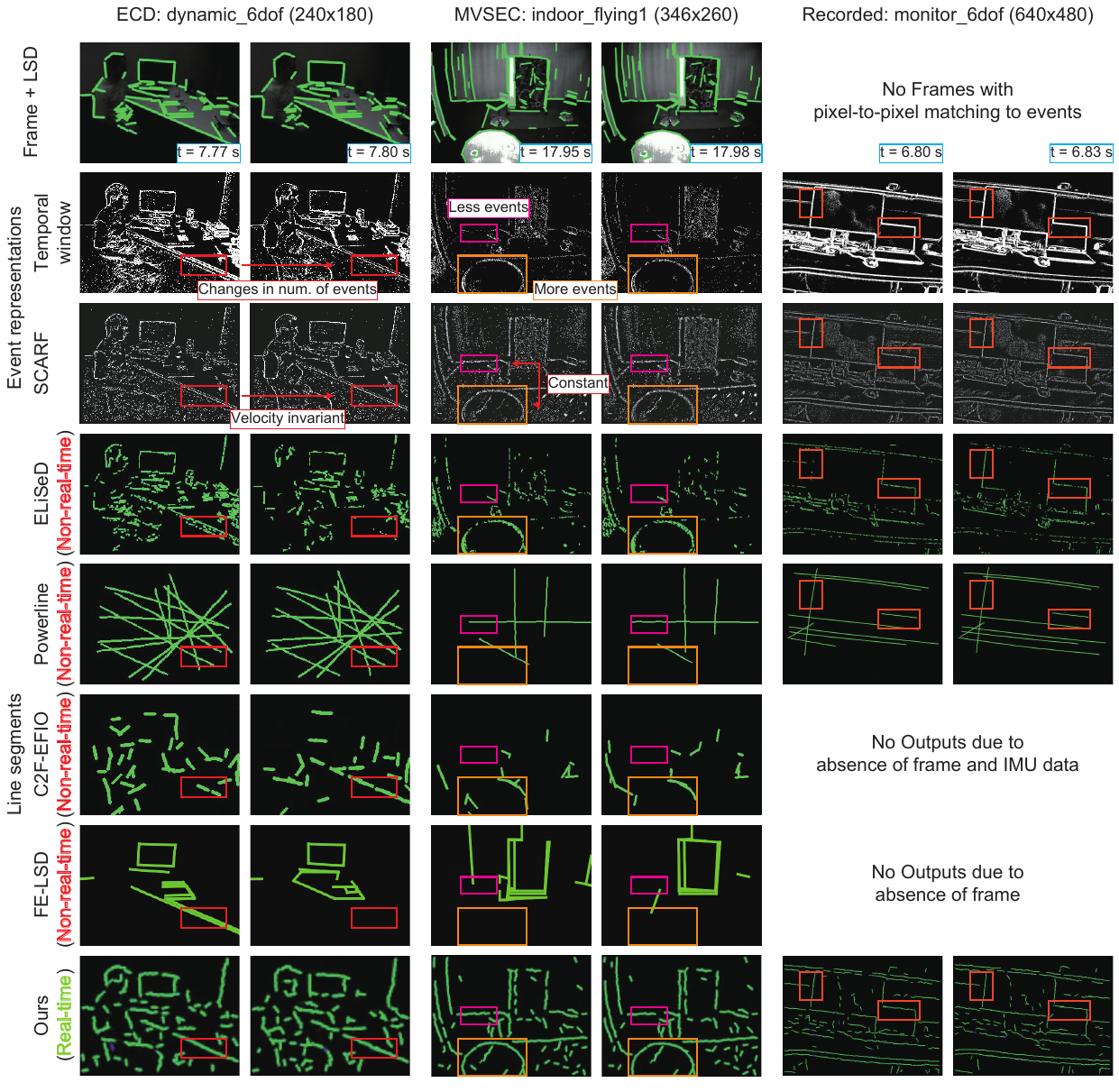}
    \caption{Qualitative comparison of line segment detection and tracking results for ELiSeD, Powerline, C2F-EFIO, FE-LSD, and our method. In the ECD dataset (dynamic\_6dof), motion changes between consecutive frames cause variations in relative velocity, which in turn lead to differences in the number of generated events. These variations are visualized as brightness differences in the temporal window, highlighted in the red square. In the MVSEC dataset (indoor\_flying1), the presence of objects at varying depths also results in differences in relative velocity and event density, as indicated by the light orange and pink squares.}
    \label{fig:line-track:4_qualitative}
\end{figure*}

We conducted a qualitative evaluation by assessing tracking accuracy with the recorded data ``monitor\_6dof'' (640x480), ECD ``dynamic\_6dof'', and MVSEC ``indoor\_flying1''. Figure~\ref{fig:line-track:4_qualitative} displays the comparison between temporal window representation~\cite{Iacono2018} and SCARF for event representation, and line segments from ELiSeD, Powerline, C2F-EFIO, FE-LSD, and ours. ELiSeD, PowerLine, C2F-EFIO, and FE-LSD required high computational cost and were not capable of real-time processing, while our method performed real-time processing. In ELiSeD, the extracted line segments were generally shorter than those from other methods, leading to visible discontinuities in the representation of object boundaries, particularly under low number of events. In contrast, Powerline produced longer line segments; however, these often failed to capture fine details or the shapes of small and curvilinear objects, as observed in the MVSEC dataset. The line segments extracted by C2F-EFIO and FE-LSD were of intermediate length, longer than those from ELiSeD and shorter than those from Powerline, providing a reasonable balance for geometric encoding.
Thanks to the lattice-based detection and tracking framework, our method extracted a larger number of line segments compared to both C2F-EFIO and FE-LSD. Furthermore, in comparison to FE-LSD, the temporal consistency of line segments between successive frames was notably improved in our method, owing to the tracking mechanism and the velocity-invariant event representation SCARF as shown in squares of Fig. \ref{fig:line-track:4_qualitative}.

\subsection{Quantitative Evaluation}
\label{sec:line-track:quantitative_eval}
\begin{figure}[t]
    \centering
    \includegraphics[width=1.0\linewidth]{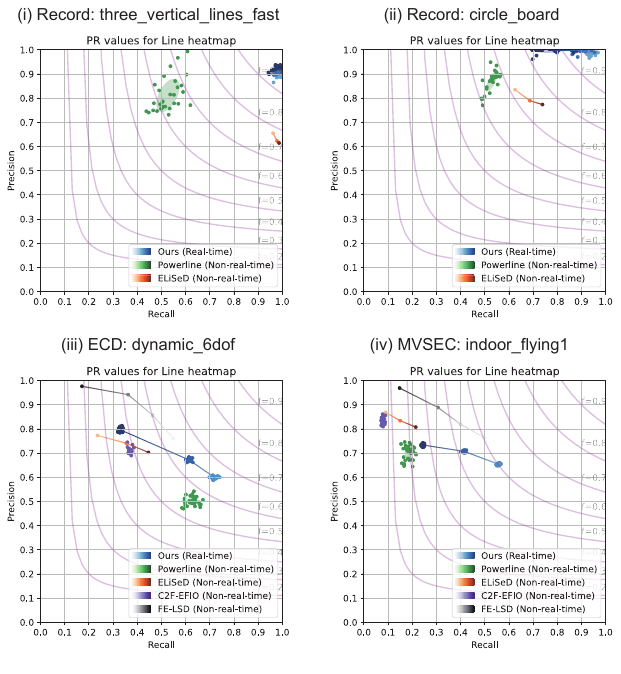}
    \caption{Precision and Recall distributions of line heat maps in four dataset and with five methods. The results of our method, ELiSeD, and FE-LSD for different thresholds are represented by using a color gradient. The purple curves show F scores as harmonic mean of the precision and recall.}
    \label{fig:line-track:4_accuracy}
\end{figure}
We quantitatively evaluate our method and baselines in terms of line segment accuracy (precision and recall), real-time performance, and parameter sensitivity (block size $b$, SCARF ratio $\alpha$, threshold $f_{th}$).
Line heat map accuracy~\cite{Zhou2019, Huang2018} was computed on two recorded datasets (640×480), ``three\_vertical\_lines\_fast'' and ``circle\_board'', and two public datasets, ECD~\cite{Mueggler2017} ``dynamic\_6dof'' (240×180) and MVSEC~\cite{MVSEC2018} ``indoor\_flying1'' (346×260). Tolerances were set to 1\% (recorded) and 2\% (public) of image diagonal~\cite{Huang2018}.
We computed precision and recall using the cumulative sum at each second over 10 seconds to evaluate temporal variations of accuracy. Figure~\ref{fig:line-track:4_accuracy} shows the distributions for Powerline, C2F-EFIO, and ours (30 trials) and deterministic ELiSeD and FE-LSD (1 trial).
In ELiSeD, some line segments have cracks due to the lack of events. In particular, in the dataset ``circle\_board'', horizontal line segments were often omitted due to limited vertical motion. Powerline showed similar omissions. In the dataset ``three\_vertical\_lines\_fast'', faster horizontal motion caused tracking failures, reducing recall in Powerline.
In addition, Powerline was not able to detect and track line segments accurately from complicated scene such as the dataset ``dynamic\_6dof'' and ``indoor\_flying1''. C2F-EFIO was sensitive to variations in event rate caused by changes in motion speed, which resulted in incomplete extraction of line segments across the entire field of view. This limitation contributed to its lower recall. FE-LSD could obtain line segments more accurately than ours. However, FE-LSD achieved accurate detection but required fine-tuning for curved shapes. Our method outperformed all baselines in F-score across datasets. Moreover, the velocity-invariant SCARF representation enabled uniform line segment extraction even in regions with insufficient events due to motion pattern (``circle\_board'') and excessive events due to fast motion (``three\_vertical\_lines\_fast''). In more complex scenes (``dynamic\_6dof'' and ``indoor\_flying1''), the lattice-based management efficiently facilitated line segment extraction.

\begin{table}[t]
    \centering
    \scalebox{0.6}{
    \begin{tabular}{@{}ccccc@{}}   
        \multicolumn{2}{c}{dynamic\_6dof (240x180, 0.49 Mev/s)} & ***-driven & Mean event rate & Mean process frequency \\ \hline
        ELiSeD &  & Event & 3.6e$^{-3}$ Mev/s $\downarrow$ & - \\ 
        Powerline &  & Event & 0.14 Mev/s $\downarrow$ & -  \\
        C2F-EFIO &  & Process & - & 32.3 Hz  \\
        FE-LSD w/ GPU &  & Process & - & 1.3 Hz  \\
        \multirow{3}{*}{Ours w/ tracking} & SCARF & Event & 26.0 Mev/s $\uparrow$ & -\\
          & Detection & Process & - & 1088 Hz \\
         & Tracking & Process & - & 2246 Hz \\ 
        \hline
        & & & & \\
        \multicolumn{2}{c}{monitor\_6dof (640x480, 1.81 Mev/s)} & ***-driven & Mean event rate & Mean process frequency \\ \hline
        ELiSeD &  & Event & 1.8e$^{-3}$ Mev/s $\downarrow$ & - \\ 
        Powerline &  & Event & 0.16 Mev/s $\downarrow$ & -  \\ 
        \multirow{3}{*}{Ours w/ tracking} & SCARF & Event & 24.7 Mev/s $\uparrow$ & -\\
          & Detection & Process & - & 209.2 Hz \\
         & Tracking & Process & - & 427.2 Hz \\ 
         \multirow{2}{*}{Ours w/o tracking} & SCARF & Event & 20.13 Mev/s $\uparrow$ & - \\
          & Detection & Process & - & 43.2 Hz \\
        \hline
    \end{tabular}
    }
    \caption{Real-time performance evaluation in three event-only and two event-frame hybrid methods with the dataset ``dynamic\_6dof'' and ``monitor\_6dof''. The upward arrow denotes real-time processing, while the downward arrow indicates a lag.}
    \label{tab:line-track:4_processtime}
\end{table}
For the real-time performance evaluation, we measured the average event rate in event-driven pipeline and the process frequency in process-driven pipeline. The recorded dataset ``monitor\_6dof'' was utilized as an input into three event-only methods, while the public dataset ``dynamic\_6dof'' which includes both events and frames was used for all baselines. The average event rate and processing frequency were computed by taking the mean over the processing performed between 5 and 10 seconds, after the system stabilized following data input. Table~\ref{tab:line-track:4_processtime} shows the results of the real-time performance. Compared to the average event rate of the dataset, the event-driven pipeline ELiSeD and Powerline exhibited lower event processing rates, indicating that these event-driven baselines do not operate in real-time. On the other hand, SCARF demonstrated a higher event processing rate than the input event rate, resulting in real-time performance. Furthermore, even with lower resolution in event data and GPU acceleration, process frequency in FE-LSD and C2F-EFIO is much lower than our method. In our method, adding tracking to the line segment detection resulted in significantly higher processing frequency in detection and tracking through lightweight processing based on endpoint perturbation, ensuring greater efficiency.

\begin{figure}[t]
    \centering
    \includegraphics[width=1.0\linewidth]{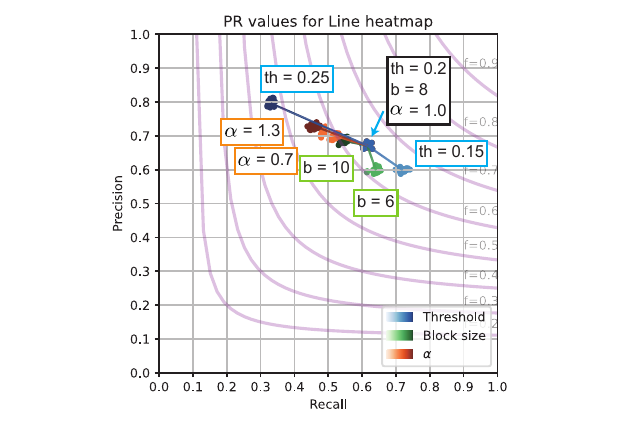}
    \caption{Sensitivity evaluation in threshold $f_{th}$, the block size $b$, and $\alpha$ in our method with respect to accuracy}
    \label{fig:line-track:4_sensitivity}
\end{figure}

Finally, we evaluated our method sensitivity to parameters, such as threshold $f_{th}$, block size $b$, and $\alpha$ for the SCARF buffer size $N$ with respect to accuracy and real-time performance. The input dataset is the first 10 seconds of ``dynamic\_6dof'' (240x180) whose average event rate is 0.49 Mev/s. As shown in Fig.~\ref{fig:line-track:4_sensitivity} and Table~\ref{tab:line-track:4_sensitivity}, we prepared 7 parameter sets and 30 trials in each parameter set for the sensitivity evaluation. These results show that a larger threshold $f_{th}$ results in higher precision and lower recall. In addition, it leads to a lower detection process frequency and a higher tracking process frequency. This result is because a higher threshold reduced the number of successful detections, decreasing the number of line segments available for tracking. As a result, the accuracy of the detected line segments improved.
\begin{table}[t]
    \centering
    \scalebox{0.6}{
    \begin{tabular}{cccccc}   
          threshold $f_{th}$ & block size $b$ & $\alpha$ & Subprocess & ***-driven & Average process speed  \\ \hline
        \multirow{3}{*}{0.2} & \multirow{3}{*}{8} & \multirow{3}{*}{1.0} &  SCARF & Event & 25.96 Mev/s \\
          & & & Detection & Process & 1087.8 Hz \\
         & & & Tracking & Process & 2246.2 Hz \\ \hline

         \multirow{3}{*}{0.15} & \multirow{3}{*}{8} & \multirow{3}{*}{1.0} &  SCARF & Event & 28.43 Mev/s \\
          & & & Detection & Process & 1581.2 Hz \\
         & & & Tracking & Process & 2116.8 Hz \\ \hline

         \multirow{3}{*}{0.25} & \multirow{3}{*}{8} & \multirow{3}{*}{1.0} &  SCARF & Event & 28.18 Mev/s \\
          & & & Detection & Process & 765.9 Hz \\
         & & & Tracking & Process & 2610.8 Hz \\ \hline

         \multirow{3}{*}{0.2} & \multirow{3}{*}{10} & \multirow{3}{*}{1.0} &  SCARF & Event & 27.18 Mev/s \\
          & & & Detection & Process & 1376.1 Hz \\
         & & & Tracking & Process & 2304.4 Hz \\ \hline

        \multirow{3}{*}{0.2} & \multirow{3}{*}{6} & \multirow{3}{*}{1.0} &  SCARF & Event & 23.69 Mev/s \\
          & & & Detection & Process & 961.8 Hz \\
         & & & Tracking & Process & 1956.2 Hz \\ \hline

        \multirow{3}{*}{0.2} & \multirow{3}{*}{8} & \multirow{3}{*}{1.3} &  SCARF & Event & 26.88 Mev/s \\
          & & & Detection & Process & 836.1 Hz \\
         & & & Tracking & Process & 1716.1 Hz \\ \hline

         \multirow{3}{*}{0.2} & \multirow{3}{*}{8} & \multirow{3}{*}{0.7} &  SCARF & Event & 23.83 Mev/s \\
          & & & Detection & Process & 1025.4 Hz \\
         & & & Tracking & Process & 3123.5 Hz \\
         
        \hline
    \end{tabular}
    }
    \caption{Sensitivity evaluation in threshold $f_{th}$, the block size $b$, and $\alpha$ in our method with respect to real-time performance}
    \label{tab:line-track:4_sensitivity}
\end{table}
Larger block size $b$ results in higher precision, lower recall, and higher process frequency for both detection and tracking. A larger block size extends line segments and reduces their total number. This improved the accuracy of each line segment and increased the process frequency. Regarding the sensitivity of $\alpha$, precision increased and recall decrease regardless of whether $\alpha$ was large or small. However, a small $\alpha$ leads to higher processing frequency for both detection and tracking. Reducing $\alpha$ decreases the number of events in the buffer, meaning that a fitting score $f$ above the threshold $f_{th}$ indicates more accurate fitting even with fewer events. On the other hand, increasing $\alpha$ amplifies the effect of the $\frac{1}{N}$ term in Eq.~(\ref{eq:line-track:3_effectiveratio}), also leading to a smaller effective event ratio $r_e$. As a result, similar to the larger $\alpha$, the line fitting becomes more precise. Besides, reducing $\alpha$ decreases the number of processed events, lowering the computational cost and significantly increasing the processing frequency for both detection and tracking.

\section{Conclusion}
We have introduced a novel lattice-allocated real-time line segment detection and tracking pipeline that relies only on an event-based camera. This consists of parallelized an event-driven subprocess to store events efficiently with velocity-invariant manner in SCARF, and two process-driven subprocesses to initialize line segments by calculating endpoints and fitting scores in detection and keep updating line states by endpoints perturbation. Throughout all experiments, the proposed pipeline maintained line segments in real-time and improved the line segment accuracy.

Future work includes clustering multiple line segments from blocks into one ``strong'' line segment to improve temporal consistency. This makes line segments more stable, enabling potential robotics applications such as SLAM.
\section*{Acknowledgments}
The authors thank Sony Interactive Entertainment for their research support and funding. 

\section{Supplementary}
\subsection{Videos}
\label{sec:video}
\begin{figure}[t]
    \centering
    \includegraphics[width=1.0\linewidth]{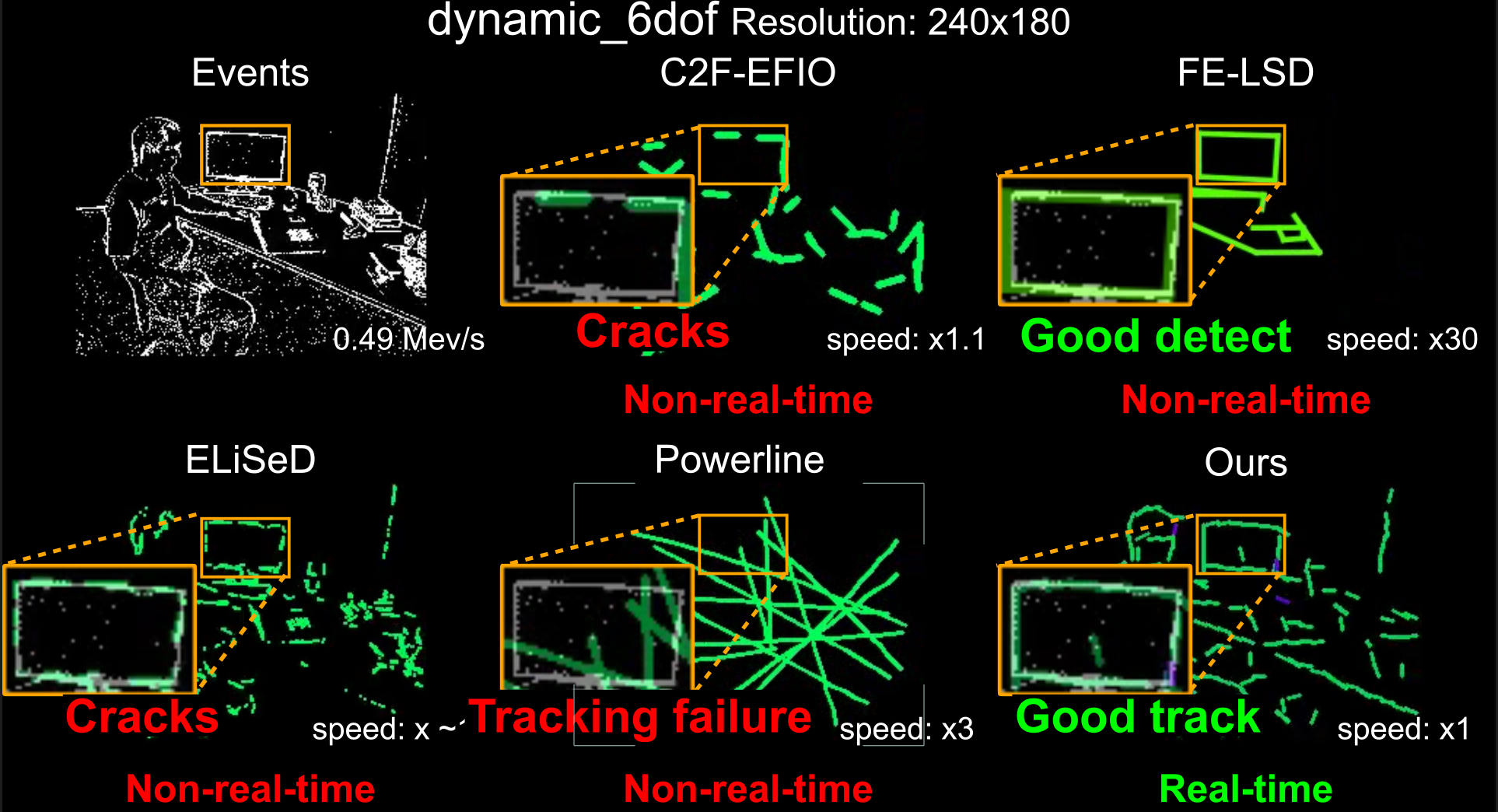}
    \caption{The part of the attached video representing line segments results with ECD dataset dynamic\_6dof from four baselines and our method. Our method outperforms others in terms of processing speed and accuracy.}
    \label{fig:line-track:X_video}
\end{figure}
We encourage readers to inspect the attached video as a supplementary material. As shown in Fig. \ref{fig:line-track:X_video}, this video summarizes line segment results from all methods with all datasets. Especially, the online demo with a modern event camera (640x480) is also presented, proving that our method works in real time in the real-world environment.

\subsection{Experimental details}
\begin{table}[t]
\centering
\scalebox{0.8}{
\begin{tabular}{@{}llc@{}}
    \toprule
    \textbf{Baselines} & \multicolumn{2}{c}{\textbf{Hyper-parameters}} \\
    \midrule
    ELiSeD~\cite{Brändli2016} & tolerance angle [deg] & 11.25, \underline{\textbf{22.5}}, 45 \\
    Powerline~\cite{Dietsche2021} & \multicolumn{2}{c}{None} \\
    C2F-EFIO~\cite{Choi2025} & \multicolumn{2}{c}{None} \\
    FE-LSD~\cite{Huai2023} & threshold & 0.15, 0.3, \underline{\textbf{0.6}}, 0.9 \\
    \midrule
    \multirow{11}{*}{Ours}
     & $\alpha$ & 0.7, \underline{\textbf{1.0}}, 1.3 \\
     & threshold $f_{th}$ & 0.15, \underline{\textbf{0.2}}, 0.25 \\
     & block size $b$ [px] & \\
     & ~~~~~~240x180: & 6, \underline{\textbf{8}}, 10 \\
     & ~~~~~~346x260: & \underline{\textbf{10}} \\
     & ~~~~~~640x480: & \underline{\textbf{14}} \\
     & perturbation size $\Delta \bm{q}$ [px] & \\
     & ~~~~~~240x180: & \underline{\textbf{0.8}} \\
     & ~~~~~~346x260: & \underline{\textbf{1.1}} \\
     & ~~~~~~640x480: & \underline{\textbf{2.5}} \\
     & distance threshold $d_{max}$ [px] & \underline{\textbf{0.2}}  $\times\,b$ \\
    \bottomrule
    \end{tabular}
    }
    \caption{Hyper-parameters from each baseline for evaluations. All bold values with underlines are default hyper-parameters for qualitative evaluation. Other values are used for quantitative evaluations, such as Precision and Recall distributions of line accuracy and sensitivity evaluations.}
    \label{tab:line-track:X_parameters}
\end{table}
Table~\ref{tab:line-track:X_parameters} summarizes the hyper-parameter settings used for each method.
Powerline and C2F-EFIO do not include tunable hyper-parameters for balancing precision and recall in line segment accuracy.
In contrast, ELiSeD, FE-LSD, and our method allow control over this trade-off via threshold values, resulting in the PR curves shown in Fig.~\ref{fig:line-track:4_accuracy}.
The units of the block size $b$, the perturbation size $\Delta \bm{q}$, and the distance threshold $d_{max}$ are in pixels; their values are scaled according to the resolution of each dataset.

\subsection{Qualitative Evaluation in Velocity-invariant Event-based Representations}
\label{sec:event-representations}
\begin{figure}[t]
    \centering
    \includegraphics[width=1.0\linewidth]{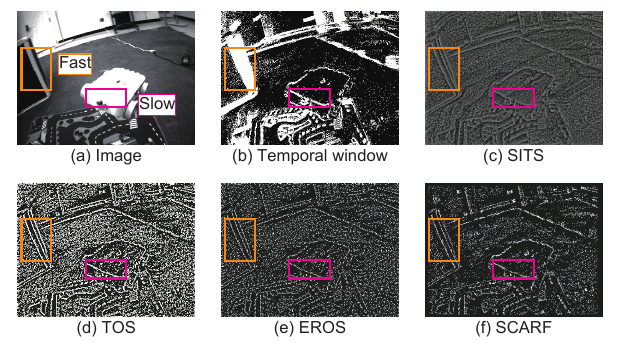}
    \caption{Qualitative comparison in velocity-invariant event-based representations (b) temporal windows \cite{Iacono2018}, (c) SITS \cite{SITS2023}, (d) TOS \cite{glover2021luvharris}, (e) EROS \cite{glover2024edopt}, and (f) SCARF with the ``boxes\_seq\_01'' in EVIMO \cite{EVIMO2019} dataset. As shown in the (a) Image, EVIMO includes multiple objects moving separately at different speed, useful for motion segmentation tasks.}
    \label{fig:line-track:X_representations}
\end{figure}
Figure~\ref{fig:line-track:X_representations} compares various velocity-invariant event-based representations. Although the temporal window contains motion blur caused by fast-moving objects, all methods mitigate this effect to some extent. However, (c) SITS, (d) TOS, and (e) EROS introduce artifacts across the entire field of view. In contrast, (f) SCARF effectively removes redundant events and preserves sharp edges of multiple objects moving at different speeds. This is achieved through the FIFO mechanism within the buffer in each block and the ``inhibitory'' role of inactive events.

\subsection{Visualization of Line Segments Results in Quantitative Evaluation}
\label{sec:detail_qualitative}
\begin{figure*}[t]
    \centering
    \includegraphics[width=1.0\linewidth]{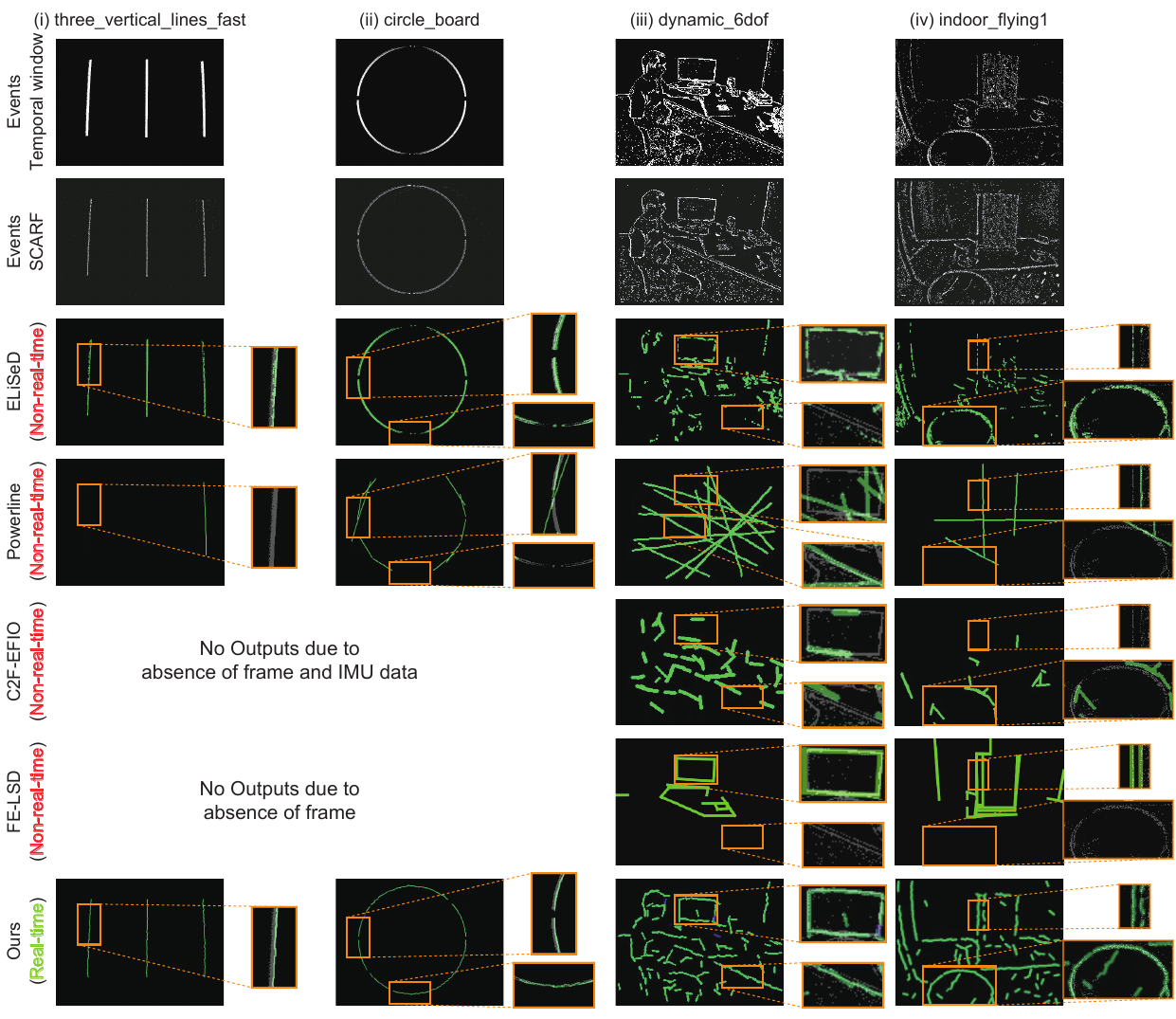}
    \caption{Detailed visualized results of line segments from all methods with datasets used in Sec. \ref{sec:line-track:quantitative_eval}. The zoomed-in images illustrate the overlay of events (gray) and the extracted line segments (green: tracked, blue: detected only in ours).
}
    \label{fig:line-track:X_visualization}
\end{figure*}
Figure \ref{fig:line-track:X_visualization} visualizes line segments from all methods with datasets used in quantitative evaluation Sec. \ref{sec:line-track:quantitative_eval}. The fast motion in the dataset ``three\_vertical\_lines\_fast'' generated a larger number of events, which in turn led to tracking failures in Powerline. In contrast, in the dataset ``circle\_board'', the vertical motion is limited, so horizontal line segments are omitted due to the lack of events. This difference in results stems from the fact that conventional storage for event data is velocity-variant. On the other hand, our method, SCARF, stores event data in a velocity-invariant manner, allowing our approach to consistently extract line segments under both fast and slow motion conditions.

\subsection{Limitations}
\label{sec:limitations}
\begin{table}[t]
    \centering
    \scalebox{0.8}{
    \begin{tabular}{@{}ccccc@{}}
         \multicolumn{2}{l}{dynamic\_6dof (240x180)} & & & \\
         \hline
         & Num. line segments & mean & std & max  \\ 
        ELiSeD & 7420 & 0.08 & 0.16 & 2.13 \\
        Powerline & 43 & 3.01 & 3.07 & 10.16 \\
        C2F-EFIO & 1425 & 0.43 & 0.56 & 3.53 \\
        Ours & 2688 & 0.97 & 1.11 & 13.32 \\
        \hline
        & & & & \\
        \multicolumn{2}{l}{monitor\_6dof (640x480)} & & & \\
         \hline
         & Num. line segments & mean & std & max  \\ 
        ELiSeD & 1815 & 0.06 & 0.08 & 0.80 \\
        Powerline & 21 & 6.69 & 5.60 & 14.85 \\ 
        Ours & 8989 & 0.40 & 0.57 & 11.25 \\
        \hline
    \end{tabular}
    }
    \caption{Lifetime of line segment extracted by four methods including tracking during the first 10 seconds with the dataset ``dynamic\_6dof''}
    \label{tab:line-track:4_lifetime}
\end{table}
\begin{figure}[t]
    \centering
    \includegraphics[width=\linewidth]{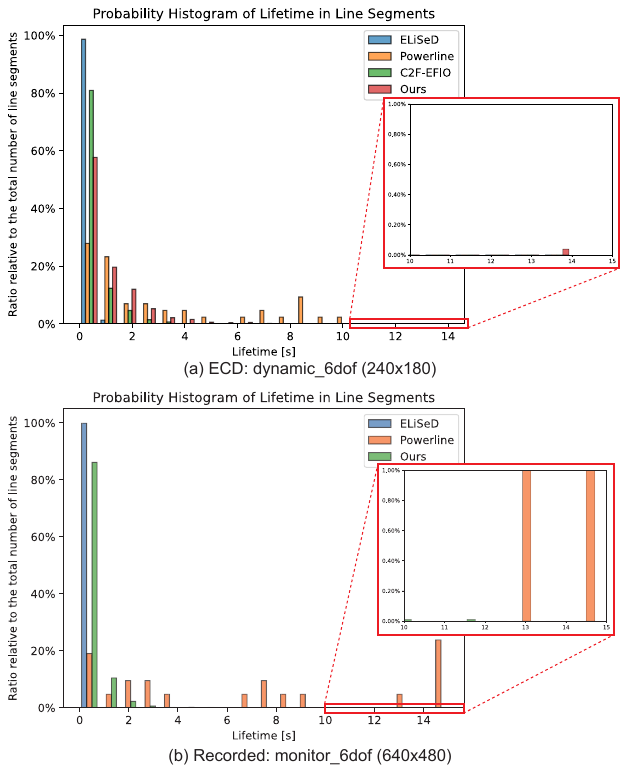}
    \caption{Probability histograms of line segment lifetime from ELiSeD, Powerline, C2F-EFIO, and ours. The zoomed-in graph demonstrates that our method successfully extracts long-lived line segments, corresponding to Powerline.}
    \label{fig:line-track:4_histogram_lifetime}
\end{figure}
The average lifetime of line segments extracted by ours is shorter than Powerline. As shown in Table~\ref{tab:line-track:4_lifetime} and Fig.~\ref{fig:line-track:4_histogram_lifetime}, to evaluate the lifetime of line segments, we computed the average and maximum duration of line segment extracted during the first 10 seconds when processing the public dataset ``dynamic\_6dof'' and the recorded dataset ``dynamic\_6dof''. ELiSeD is based on pixel-wise non-parametric approach, meaning that the line segments become short-lived. Even the maximum lifetime of line segments in ELiSeD was about 2 seconds in ``dynamic\_6dof'' and shorter than 1 second in ``monitor\_6dof''. On the other hand, Powerline enabled to keep long-lived line segments due to spatio-temporal based approach. Compared to the three baselines, our method extracted a larger number of line segments, resulting in a longer average lifetime than ELiSeD and C2F-EFIO but shorter than Powerline. However, on the benefit of tracking, the same line segment ID could be maintained for a longer duration, bringing the maximum lifetime longer than C2F-EFIO and closer to Powerline. Furthermore, clustering multiple line segments from blocks into one ``strong'' line segment would improve temporal consistency.

{
    \small
    \bibliographystyle{ieeenat_fullname}
    \bibliography{main}

\begin{thebibliography}{36}
\providecommand{\natexlab}[1]{#1}
\providecommand{\url}[1]{\texttt{#1}}
\expandafter\ifx\csname urlstyle\endcsname\relax
  \providecommand{\doi}[1]{doi: #1}\else
  \providecommand{\doi}{doi: \begingroup \urlstyle{rm}\Url}\fi

\bibitem[Acin et~al.(2023)Acin, Jacob, Simon-Chane, and Histace]{SITS2023}
Laure Acin, Pierre Jacob, Camille Simon-Chane, and Aymeric Histace.
\newblock {VK-SITS}: a robust time-surface for fast event-based recognition.
\newblock In \emph{2023 Twelfth International Conference on Image Processing Theory, Tools and Applications (IPTA)}, pages 1--6, 2023.

\bibitem[Alzugaray and Chli(2020)]{alzugaray2020}
I. Alzugaray and M. Chli.
\newblock {HASTE}: multi-hypothesis asynchronous speeded-up tracking of events.
\newblock In \emph{31st British Machine Vision Virtual Conference (BMVC)}, 2020.

\bibitem[Benosman et~al.(2012)Benosman, Ieng, Clercq, Bartolozzi, and Srinivasan]{Benosman2012}
Ryad Benosman, Sio-Hoi Ieng, Charles Clercq, Chiara Bartolozzi, and Mandyam Srinivasan.
\newblock Asynchronous frameless event-based optical flow.
\newblock \emph{Neural Networks}, 27:\penalty0 32--37, 2012.

\bibitem[Brändli et~al.(2016)Brändli, Strubel, Keller, Scaramuzza, and Delbruck]{Brändli2016}
Christian Brändli, Jonas Strubel, Susanne Keller, Davide Scaramuzza, and Tobi Delbruck.
\newblock {ELiSeD} — an event-based line segment detector.
\newblock In \emph{2016 Second International Conference on Event-based Control, Communication, and Signal Processing (EBCCSP)}, pages 1--7, 2016.

\bibitem[Chakravarthi et~al.(2024)Chakravarthi, Verma, Daniilidis, Fermuller, and Yang]{Bharatesh2024}
Bharatesh Chakravarthi, Aayush~Atul Verma, Kostas Daniilidis, Cornelia Fermuller, and Yezhou Yang.
\newblock Recent event camera innovations: A survey.
\newblock Presented at NeVi 2024, Workshop on Neuromorphic Vision: Advantages and Applications of Event Cameras, ECCV2024, 2024.

\bibitem[Chamorro et~al.(2022)Chamorro, Solà, and Andrade-Cetto]{Chamorro2022}
William Chamorro, Joan Solà, and Juan Andrade-Cetto.
\newblock Event-based line slam in real-time.
\newblock \emph{IEEE Robotics and Automation Letters}, 7\penalty0 (3):\penalty0 8146--8153, 2022.

\bibitem[Choi et~al.(2025)Choi, Lee, and Gook~Park]{Choi2025}
Byeongpil Choi, Hanyeol Lee, and Chan Gook~Park.
\newblock Event-frame-inertial odometry using point and line features based on coarse-to-fine motion compensation.
\newblock \emph{IEEE Robotics and Automation Letters}, 10\penalty0 (3):\penalty0 2622--2629, 2025.

\bibitem[Denis et~al.(2008)Denis, Elder, and Estrada]{Denis2008}
Patrick Denis, James~H. Elder, and Francisco~J. Estrada.
\newblock Efficient edge-based methods for estimating manhattan frames in urban imagery.
\newblock In \emph{Computer Vision -- ECCV 2008}, pages 197--210, Berlin, Heidelberg, 2008. Springer Berlin Heidelberg.

\bibitem[Dietsche et~al.(2021)Dietsche, Cioffi, Hidalgo-Carrio, and Scaramuzza]{Dietsche2021}
Alexander Dietsche, Giovanni Cioffi, Javier Hidalgo-Carrio, and Davide Scaramuzza.
\newblock Powerline tracking with event cameras.
\newblock In \emph{IEEE/RSJ Int. Conf. Intell. Robot. Syst. (IROS)}, 2021.

\bibitem[Dimitrova et~al.(2020)Dimitrova, Gehrig, Brescianini, and Scaramuzza]{Dimitrova2020}
Rika~Sugimoto Dimitrova, Mathias Gehrig, Dario Brescianini, and Davide Scaramuzza.
\newblock Towards low-latency high-bandwidth control of quadrotors using event cameras.
\newblock In \emph{2020 IEEE International Conference on Robotics and Automation (ICRA)}, pages 4294--4300, 2020.

\bibitem[Everding and Conradt(2018)]{Everding2018}
Lukas Everding and Jörg Conradt.
\newblock Low-latency line tracking using event-based dynamic vision sensors.
\newblock \emph{Frontiers in Neurorobotics}, 12, 2018.

\bibitem[Gallego et~al.(2018)Gallego, Rebecq, and Scaramuzza]{gallego2018unifying}
Guillermo Gallego, Henri Rebecq, and Davide Scaramuzza.
\newblock A unifying contrast maximization framework for event cameras, with applications to motion, depth, and optical flow estimation.
\newblock In \emph{Proceedings of the IEEE conference on computer vision and pattern recognition}, pages 3867--3876, 2018.

\bibitem[Gallego et~al.(2022)Gallego, Delbrück, Orchard, Bartolozzi, Taba, Censi, Leutenegger, Davison, Conradt, Daniilidis, and Scaramuzza]{Gallego2022}
Guillermo Gallego, Tobi Delbrück, Garrick Orchard, Chiara Bartolozzi, Brian Taba, Andrea Censi, Stefan Leutenegger, Andrew~J. Davison, Jörg Conradt, Kostas Daniilidis, and Davide Scaramuzza.
\newblock Event-based vision: A survey.
\newblock \emph{IEEE Transactions on Pattern Analysis and Machine Intelligence}, 44\penalty0 (1):\penalty0 154--180, 2022.

\bibitem[Glover et~al.(2021)Glover, Dinale, Rosa, Bamford, and Bartolozzi]{glover2021luvharris}
Arren Glover, Aiko Dinale, Leandro De~Souza Rosa, Simeon Bamford, and Chiara Bartolozzi.
\newblock {luvharris}: A practical corner detector for event-cameras.
\newblock \emph{IEEE Transactions on Pattern Analysis and Machine Intelligence}, 44\penalty0 (12):\penalty0 10087--10098, 2021.

\bibitem[Glover et~al.(2024)Glover, Gava, Li, and Bartolozzi]{glover2024edopt}
Arren Glover, Luna Gava, Zhichao Li, and Chiara Bartolozzi.
\newblock {EDOPT}: Event-camera 6-dof dynamic object pose tracking.
\newblock In \emph{2024 IEEE International Conference on Robotics and Automation (ICRA)}, pages 18200--18206. IEEE, 2024.

\bibitem[Grompone~von Gioi et~al.(2012)Grompone~von Gioi, Jakubowicz, Morel, and Randall]{Grompone2012}
Rafael Grompone~von Gioi, Jérémie Jakubowicz, Jean-Michel Morel, and Gregory Randall.
\newblock {{LSD}: a Line Segment Detector}.
\newblock \emph{{Image Processing On Line}}, 2:\penalty0 35--55, 2012.

\bibitem[Gu et~al.(2022)Gu, Ko, Go, Lee, Lee, and Shin]{Gu2022}
Geonmo Gu, Byungsoo Ko, SeoungHyun Go, Sung-Hyun Lee, Jingeun Lee, and Minchul Shin.
\newblock Towards light-weight and real-time line segment detection.
\newblock \emph{Proceedings of the AAAI Conference on Artificial Intelligence}, 36\penalty0 (1):\penalty0 726--734, 2022.

\bibitem[Gómez~Eguíluz et~al.(2020)Gómez~Eguíluz, Rodríguez-Gómez, Martínez-de Dios, and Ollero]{Gómez2020}
A. Gómez~Eguíluz, J.P. Rodríguez-Gómez, J.R. Martínez-de Dios, and A. Ollero.
\newblock Asynchronous event-based line tracking for time-to-contact maneuvers in uas.
\newblock In \emph{2020 IEEE/RSJ International Conference on Intelligent Robots and Systems (IROS)}, pages 5978--5985, 2020.

\bibitem[Huang et~al.(2018)Huang, Wang, Zhou, Ding, Gao, and Ma]{Huang2018}
Kun Huang, Yifan Wang, Zihan Zhou, Tianjiao Ding, Shenghua Gao, and Yi Ma.
\newblock Learning to parse wireframes in images of man-made environments.
\newblock In \emph{2018 IEEE/CVF Conference on Computer Vision and Pattern Recognition}, pages 626--635, 2018.

\bibitem[Iacono et~al.(2018)Iacono, Weber, Glover, and Bartolozzi]{Iacono2018}
Massimiliano Iacono, Stefan Weber, Arren Glover, and Chiara Bartolozzi.
\newblock Towards event-driven object detection with off-the-shelf deep learning.
\newblock In \emph{2018 IEEE/RSJ International Conference on Intelligent Robots and Systems (IROS)}, pages 1--9, 2018.

\bibitem[Ikura et~al.(2024)Ikura, Gentil, Müller, Schuler, Yamashita, and Stürzl]{Ikura2024}
Mikihiro Ikura, Cedric~Le Gentil, Marcus~G. Müller, Florian Schuler, Atsushi Yamashita, and Wolfgang Stürzl.
\newblock {RATE}: Real-time asynchronous feature tracking with event cameras.
\newblock In \emph{2024 IEEE/RSJ International Conference on Intelligent Robots and Systems (IROS)}, pages 11662--11669, 2024.

\bibitem[Le~Gentil et~al.(2020)Le~Gentil, Tschopp, Alzugaray, Vidal-Calleja, Siegwart, and Nieto]{Gentil2020}
Cedric Le~Gentil, Florian Tschopp, Ignacio Alzugaray, Teresa Vidal-Calleja, Roland Siegwart, and Juan Nieto.
\newblock {IDOL}: A framework for imu-dvs odometry using lines.
\newblock In \emph{2020 IEEE/RSJ International Conference on Intelligent Robots and Systems (IROS)}, pages 5863--5870, 2020.

\bibitem[Liu et~al.(2024)Liu, Guan, Shang, Yu, and Kneip]{Zibin2024}
Zibin Liu, Banglei Guan, Yang Shang, Qifeng Yu, and Laurent Kneip.
\newblock Line-based 6-dof object pose estimation and tracking with an event camera.
\newblock \emph{IEEE Transactions on Image Processing}, 33:\penalty0 4765--4780, 2024.

\bibitem[Liu et~al.(2025)Liu, Guan, Shang, Bian, Sun, and Yu]{Zibin2025}
Zibin Liu, Banglei Guan, Yang Shang, Yifei Bian, Pengju Sun, and Qifeng Yu.
\newblock Stereo event-based, 6-dof pose tracking for uncooperative spacecraft.
\newblock \emph{IEEE Transactions on Geoscience and Remote Sensing}, 63:\penalty0 1--13, 2025.

\bibitem[Manderscheid et~al.(2019)Manderscheid, Sironi, Bourdis, Migliore, and Lepetit]{manderscheid2019speed}
Jacques Manderscheid, Amos Sironi, Nicolas Bourdis, Davide Migliore, and Vincent Lepetit.
\newblock Speed invariant time surface for learning to detect corner points with event-based cameras.
\newblock In \emph{Proceedings of the IEEE/CVF Conference on Computer Vision and Pattern Recognition}, pages 10245--10254, 2019.

\bibitem[Micusik and Wildenauer(2014)]{Micusik2014}
Branislav Micusik and Horst Wildenauer.
\newblock Structure from motion with line segments under relaxed endpoint constraints.
\newblock In \emph{2014 2nd International Conference on 3D Vision}, pages 13--19, 2014.

\bibitem[Mitrokhin et~al.(2019)Mitrokhin, Ye, Fermüller, Aloimonos, and Delbruck]{EVIMO2019}
Anton Mitrokhin, Chengxi Ye, Cornelia Fermüller, Yiannis Aloimonos, and Tobi Delbruck.
\newblock {EV-IMO}: Motion segmentation dataset and learning pipeline for event cameras.
\newblock In \emph{2019 IEEE/RSJ International Conference on Intelligent Robots and Systems (IROS)}, pages 6105--6112, 2019.

\bibitem[Mueggler et~al.(2017{\natexlab{a}})Mueggler, Bartolozzi, and Scaramuzza]{Mueggler17BMVC}
Elias Mueggler, Chiara Bartolozzi, and Davide Scaramuzza.
\newblock Fast event-based corner detection.
\newblock In \emph{British Machine Vision Conference (BMVC)}, 2017{\natexlab{a}}.

\bibitem[Mueggler et~al.(2017{\natexlab{b}})Mueggler, Rebecq, Gallego, Delbruck, and Scaramuzza]{Mueggler2017}
Elias Mueggler, Henri Rebecq, Guillermo Gallego, Tobi Delbruck, and Davide Scaramuzza.
\newblock The event-camera dataset and simulator: Event-based data for pose estimation, visual odometry, and slam.
\newblock \emph{The International Journal of Robotics Research}, 36\penalty0 (2):\penalty0 142–149, 2017{\natexlab{b}}.

\bibitem[Pautrat et~al.(2023)Pautrat, Barath, Larsson, Oswald, and Pollefeys]{Pautrat2023}
R\'emi Pautrat, Daniel Barath, Viktor Larsson, Martin~R. Oswald, and Marc Pollefeys.
\newblock {DeepLSD}: Line segment detection and refinement with deep image gradients.
\newblock In \emph{Proceedings of the IEEE/CVF Conference on Computer Vision and Pattern Recognition (CVPR)}, pages 17327--17336, 2023.

\bibitem[Qiao et~al.(2021)Qiao, Bai, Xiang, Qian, and Bi]{Qiao2021}
Chengyu Qiao, Tingming Bai, Zhiyu Xiang, Qi Qian, and Yunfeng Bi.
\newblock {Superline}: A robust line segment feature for visual slam.
\newblock In \emph{2021 IEEE/RSJ International Conference on Intelligent Robots and Systems (IROS)}, pages 5664--5670, 2021.

\bibitem[Wang et~al.(2024)Wang, Zhang, Yu, and Wan]{Wang2024}
Xinya Wang, Haitian Zhang, Huai Yu, and Xianrong Wan.
\newblock {EvLSD-IED}: Event-based line segment detection with image-to-event distillation.
\newblock \emph{IEEE Transactions on Instrumentation and Measurement}, 73:\penalty0 1--12, 2024.

\bibitem[Xu et~al.(2017)Xu, Zhang, Cheng, and Koch]{Xu2017}
Chi Xu, Lilian Zhang, Li Cheng, and Reinhard Koch.
\newblock Pose estimation from line correspondences: A complete analysis and a series of solutions.
\newblock \emph{IEEE Transactions on Pattern Analysis and Machine Intelligence}, 39\penalty0 (6):\penalty0 1209--1222, 2017.

\bibitem[Yu et~al.(2023)Yu, Li, Yang, Yu, and Xia]{Huai2023}
Huai Yu, Hao Li, Wen Yang, Lei Yu, and Gui-Song Xia.
\newblock Detecting line segments in motion-blurred images with events.
\newblock \emph{IEEE Transactions on Pattern Analysis and Machine Intelligence}, pages 1--16, 2023.

\bibitem[Zhou et~al.(2019)Zhou, Qi, and Ma]{Zhou2019}
Yichao Zhou, Haozhi Qi, and Yi Ma.
\newblock End-to-end wireframe parsing.
\newblock In \emph{2019 IEEE/CVF International Conference on Computer Vision (ICCV)}, pages 962--971, 2019.

\bibitem[Zhu et~al.(2018)Zhu, Thakur, Özaslan, Pfrommer, Kumar, and Daniilidis]{MVSEC2018}
Alex~Zihao Zhu, Dinesh Thakur, Tolga Özaslan, Bernd Pfrommer, Vijay Kumar, and Kostas Daniilidis.
\newblock The multivehicle stereo event camera dataset: An event camera dataset for 3d perception.
\newblock \emph{IEEE Robotics and Automation Letters}, 3\penalty0 (3):\penalty0 2032--2039, 2018.

\end{thebibliography}
}


\end{document}